%% file: FlexibleFitnessCommunityDetection.tex
\documentclass{llncs}
\pdfoutput=1

\usepackage[pdftex]{graphicx}
\graphicspath{{./pdf/}{../jpeg/}{../eps/}{./Figures/}}
\DeclareGraphicsExtensions{.pdf,.jpeg,.png,.eps}

\usepackage[cmex10]{amsmath}
\usepackage[caption=false,font=footnotesize]{subfig}
\usepackage[ruled,vlined]{algorithm2e}
\usepackage{url}
\usepackage{multirow}

\begin{document}

\title{A Flexible Fitness Function for Community Detection in Complex Networks}

\author{Fabr\'{i}cio Olivetti de Fran\c{c}a\inst{1} \and Guilherme Palermo Coelho\inst{2}}
\institute{CMCC, Federal University of ABC (UFABC) -- Brazil \\
 \email{folivetti@ufabc.edu.br}
\and School of Technology (FT), University of Campinas (Unicamp) -- Brazil \\
 \email{guilherme@ft.unicamp.br}}

\maketitle

\begin{abstract} 
Most community detection algorithms from the literature work as optimization tools that minimize a given \textit{fitness function}, while assuming that each node belongs to a single community. Since there is no hard concept of what a community is, most proposed fitness functions focus on a particular definition. As such, these functions do not always lead to partitions that correspond to those observed in practice. This paper proposes a new flexible fitness function that allows the identification of communities with distinct characteristics. Such flexibility was evaluated through the adoption of an immune-inspired optimization algorithm, named cob-aiNet[C], to identify both disjoint and overlapping communities in a set of benchmark networks. The results have shown that the obtained partitions are much closer to the ground-truth than those obtained by the optimization of the modularity function.

\end{abstract}

\section{Introduction}
\label{sec:introduction}
\vspace{-4pt}

Complex problems from a wide range of fields can be modeled and described as complex networks~\cite{NewmanGirvan2004}. In such networks, nodes that present similar properties often tend to be linked to each other, thus forming consistent subgraphs with dense interconnections that are called \textit{communities}~\cite{NewmanGirvan2004}. The detection of communities in complex networks is an important step in the multitude of possible analyses that can be performed to such models. When a complex problem is modeled as a network, the identification of communities may allow both the comprehension of characteristics that are specific to subgroups of nodes and of how such nodes interact to each other~\cite{NewmanGirvan2004}.

Although communities play an important role in the network analysis, as they allow the identification of functional properties of a group of nodes and also of the ways complex behaviors emerge from simple individual functions, the formal definition of communities is still vague in the literature. Therefore, one of the challenges associated with the development of community detection algorithms is to select which metric should be used to properly evaluate whether a given set of nodes actually represent a community with characteristics that are relevant to the context of the problem. Such metrics become even more important if it is considered that a large part of algorithms for community detection are based on the optimization of \textit{quality functions} (or \textit{fitness} functions). 

From all quality metrics described in the literature, one of the most adopted is Modularity~\cite{NewmanGirvan2004}, which assumes that a community is a \textit{module of the network} and that two nodes belonging to the same community tend to have much higher probability of being connected to each other than that of two nodes belonging to different communities.

Given that each quality function may be intended to identify a set of communities according to one of the different existing definitions, the resulting partition of the original complex network invariably reflects the characteristics of such definitions. Besides, it is known that the optimization of some of these quality functions may lead to partitions of the network that do not correspond to the real partition observed in practice~\cite{lancichinettiEtAl20011}.

The above scenario is even worse when \textit{overlapping communities} are considered. Differently from disjoint communities of complex networks, in which each node belongs to a single community, the partition of the network into overlapping communities allows some nodes (known as \textit{bridges}) to belong to different communities at the same time. In this context, aspects such as the \textit{clustering coefficient}\footnote{Clustering coefficient corresponds to the ratio between the number of triangles formed by a given node and all possible triangles that could be formed.} of the network and the number of edges connecting a given node to its neighbors (both in and out of its community) may have different impacts when choosing a bridge in complex networks that model different real-world situations.

Therefore, in this paper a new flexible fitness function for community detection is proposed. This new metric, named \textit{Flex}, allows the user to predefine which characteristics should be present in the communities that will be obtained by the optimization process, thus allowing the identification of distinct sets of communities for the same complex network by simply adjusting a few intuitive parameters. Flex was combined with an adapted version of the immune-inspired optimization algorithm named cob-aiNet[C]~\cite{CoelhoEtAl2011} and applied to identify both disjoint and overlapping communities in a set of eight artificial and four real-world complex networks. The obtained results have shown that the partitions obtained with the optimization of this new metric are more coherent with the known real partitions than those obtained with the optimization of modularity.

This paper is organized as follows. The new flexible objective function for community detection will be presented in Section~\ref{sec:flex}, together with some insights about how it can be applied to identify overlapping communities and a brief description of the optimization algorithm adopted here. The experimental methodology and the obtained results will be discussed in Section~\ref{sec:experiments}. Finally, concluding remarks and indications for future work will be given in Section~\ref{sec:conclusion}.

\section{A Flexible Objective Function for Community Detection}
\label{sec:flex}
\vspace{-4pt}

In a broader definition, a community structure of a network is a partition of the nodes so that each partition is densely connected. The modularity metric~\cite{NewmanGirvan2004} tries to capture this definition by analyzing the difference between the number of edges inside a community and the expected number of edges that would be observed if this community was formed in a random network. Although modularity is widely adopted by the complex network community, its structure may lead to the false assumption that the number of edges between two groups decreases as the network size increases. Therefore, for larger networks, a simple connection between two nodes of different communities may result in the merging of these two communities, in order to increase (maximize) modularity. This aspect is known as the \textit{resolution limit} of the metric~\cite{fortunato2007resolution}.

Additionally, in a situation in which a given node has few links connecting it to a small community and most of its links connecting it to a large community, the optimization of modularity will often include such node into the larger community, without considering the local contribution of this node to the smaller communitiy. This can be a drawback if the node has a higher clustering coefficient with respect to the smaller community than to the larger one.

With that in mind, a new quality function for community detection, hereby called \textit{Flex}, is proposed. The optimization of Flex tries to balance two objectives at the same time: maximize both the number of links inside a community and the local clustering coefficient of each community. Additionally, it also penalizes the occurrence of open triangles (i.e., it minimizes the \textit{random model} effect~\cite{erdds1959random}).

The first step to calculate Flex for a given partition of the network is to define the \textit{Local Contribution} of a node $i$ to a given community $c$:

\begin{equation}
LC(i,c) = \alpha*\bigtriangleup(i,c) + (1-\alpha)*N(i,c) - \beta*\wedge(i,c),
\label{eq:LC}
\end{equation}

\noindent where $\bigtriangleup(i,c)$ is the ratio between the transitivity of node $i$ (number of triangles that $i$ forms) inside community $c$ and the total transitivity of this node in the full network, $N(i,c)$ is the ratio between the number of neighbors node $i$ has inside community $c$ and its total number of neighbors, and $\wedge(i,c)$ is the ratio between the number of open triangles in community $c$ that contain node $i$ and the total participation of $i$ in the whole network. Variables $\alpha$ and $\beta$ are weights that balance the importance of each term.

The weight parameter $\alpha$ directly dictates whether the optimization process will tend to insert a given node into a clustered community or into a community that contains the majority of this node's neighbors. It is important to consider both transitivity and the neighborhood of each node in the optimization, as both concepts are not necessarily related (i.e. a given node will not necessarily have high transitivity with the majority of its neighbors). Therefore, by weighting these two criteria, the user can emphasize each of them according to what is desirable in a given practical situation.

Given the Local Contribution of all nodes to each community, the \textit{Community Contribution} (CC) of a community $c$ in a given partition is defined as:

\begin{equation}
CC(c) = \sum_{i \in c}{LC(i,c)} - \frac{|c|}{|V|}^{\gamma},
\label{eq:CC}
\end{equation}

\noindent where $|.|$ is the number of elements in a set, $V$ is the set of nodes of a network and $\gamma$ is the penalization weight. The penalization in this equation is devised to avoid the generation of a trivial solution, in which the entire network forms a single community. Finally, the Flex value of a given partition $p$ is given by:

\begin{equation}
Flex(p) = \frac{1}{|V|}\sum_{c \in p}{CC(c)}
\end{equation}

It is also important to highlight that the penalization term of Eq.~\ref{eq:LC} ensures that the convergence to the random model is penalized, even if $\alpha$ favors only the number of neighbors (i.e., $\alpha=0$) in the definition of the communities.

\vspace{-4pt}
\subsection{\textbf{Applying Flex to Identify Overlapping Nodes}}
\label{sec:overlap}

The Flex fitness metric also provides insights about overlapping nodes. As the importance of transitivity and neighborhood is balanced by $\alpha$, this can be exploited to infer whether a given node also belongs to another community.

When using Flex as a fitness function, some nodes may be more sensible than others regarding the weight $\alpha$. This happens when, for example, a node has a fraction of its neighbors on a clustered community and the remaining neighbors spread across one or more communities with lower transitivity.

Therefore, a simple heuristic that allows the identification of overlapping nodes is to search for nodes that do not make significant contribution to one of the $\alpha$-weighted factors, i.e., nodes that are sensible to changes of $\alpha$. After finding those nodes, we can allocate them to other communities that share a certain fraction of neighbors with them. This heuristic is summarized in Alg.~\ref{alg:overlapheuristic}.
\vspace{-10pt}

\begin{algorithm}
\KwData{thresholds $thr\bigtriangleup$ and $thrN$ for the contribution to transitivity and neighborhood, respectively, and threshold $thrSh$ of shared neighbors between communities.}
\KwResult{New set of communities with overlapping nodes.}
\BlankLine
\For{each node $i$}{
	$c$ = community that contains $i$ \\
	\If{$\bigtriangleup(i,c) < thr\bigtriangleup$ or $N(i,c) < thrN$}{
		\For{$\forall c_j \neq c$}{
			\If{$N(i,c_j) > thrSh$}{ 
				Add $i$ to community $c_j$
			}
		}
	}
}
\label{alg:overlapheuristic}
\caption{Heuristic to find overlapping nodes.}
\end{algorithm}

\vspace{-20pt}
\subsection{The cob-aiNet[C] Algorithm}
\label{sec:cobAiNet}

As previously mentioned, an adaptation of the cob-aiNet[C] algorithm (\textit{Con\-cen\-tra\-tion-based Artificial Immune Network for Combinatorial Optimization} -- \cite{CoelhoEtAl2011}) was adopted in this paper to obtain a set of communities for complex networks that maximize the new proposed quality function (Flex).

The cob-aiNet[C] algorithm, which was originally proposed to solve combinatorial optimization problems~\cite{CoelhoEtAl2011}, was previously adapted to identify both disjoint and overlapping communities in complex networks~\cite{olivetti2013identifying}. As most of the adaptations proposed in~\cite{olivetti2013identifying} were adopted here as well, only a brief explanation of the general aspects of cob-aiNet[C] will be presented here, together with details about those aspects that differ from the adaptation proposed in~\cite{olivetti2013identifying}. For further details, the reader is referred to~\cite{CoelhoEtAl2011,olivetti2013identifying}.

The cob-aiNet[C] is a bioinspired search-based optimization algorithm that contains operators inspired in the natural immune system of vertebrates. Therefore, it evolves a population of candidate solutions of the problem (cells or possible partitions of the complex network), through a sequence of cloning, mutation and selection steps, guided by the fitness of each individual solution. 

Besides these evolutionary steps, all the cells in cob-aiNet[C] population are compared to each other and, whenever a given cell is more similar to a better one than a given threshold, its concentration (a real value assigned to each cell) is reduced. This concentration can also be increased according to the fitness of the cell (higher fitness leads to higher concentration). Such concentration-based mechanism is an essential feature of the algorithm, as it controls the number clones that will be generated for each cell at each iteration, the intensity of the mutation process that will be applied to each clone and when a given cell should be eliminated from the population (when its concentration becomes null).

When compared to the adaptations made in~\cite{olivetti2013identifying}, the only differences are associated with the new hypermutation operator, which will be discussed next, and the new approach to obtain overlapping communities described in Sect.~\ref{sec:overlap}.

\vspace{-10pt}
\subsubsection{\textbf{The new Hypermutation Operator}}
\label{sec:cobHypermutation}

To properly explain the new hypermutation operator, it is important to know that each cell in the population of the algorithm is represented as an array of integers with length equal to the number $N$ of nodes of the complex network. Each position $i$ of the array corresponds to a node of the network and assumes value $j \in \{1, 2, \ldots, N\}$ that indicates that nodes $i$ and $j$ belong to the same community. 

This operator, which is applied to all cells in the population at a given iteration, is basically a random modification of the integer values in $n_{mut}$ positions of the array (cell), being $n_{mut}$ inversely proportional to the fitness and concentration of each cell (further details can be found in~\cite{CoelhoEtAl2011,olivetti2013identifying}).

The $n_{mut}$ positions of the cell that will suffer mutation are randomly selected, and so are the values that will be inserted into these positions. However, the probability that a given value $k$ (associated with node $k$) replaces the current value in position $i$ is directly proportional to $|N(i) \cap N(k)|$, where $N(i)$ is the set of nodes that are neighbors of $i$.

\vspace{-4pt}
\section{Experimental Results}
\label{sec:experiments}

In order to assess whether Flex is able to lead to improvements in community detection, when compared to Modularity, an extensive experimental setup, composed of $8$ artificial and $4$ real world networks was devised. The artificial networks, which were generated by the toolbox provided by Lancichinetti~\cite{lancichinetti2008benchmark}, are composed by $4$ networks formed by high density communities (i.e. with a high number of internal edges), which facilitates the identification of the optimal partition, and $4$ networks with noisy communities (i.e. with higher probability of presenting edges connecting them to other communities).

The artificial networks were generated with $50$, $100$, $200$ and $500$ nodes, being these nodes with average degree of $10$ and maximum degree of $15$. Such networks were generated with a maximum of $10$ communities, $3$ overlapping nodes belonging to an average of $2$ communities and average clustering coefficient of $0.7$. The mixing parameter, which introduces noise to the network structure, was set as $0.1$ for the first set of networks (labeled Network ${50-500}$ in the tables that follow) and $0.3$ for the second set (labeled Noise Network ${50-500}$).

The real world networks (with known partitions) that were chosen for the experiments were: Zachary's Karate Club~\cite{Zachary1977}; Dolphins Social Network~\cite{Lusseau2003}; American College Football~\cite{evans2010clique}; and a network of co-purchasing of books about US politics compiled by Krebs~\cite{krebs2004social}.

A total of $20$ repetitions were performed for each network and for each fitness function adopted here (Flex and Modularity). The cob-aiNet[C] algorithm was empirically adjusted with the following parameters for all of the experiments: $\sigma_S=0.2$, maximum number of iterations equal to $1,500$, $\alpha^{Ini}=10$, $\alpha^{End}=1$, initial population with $4$ candidate solutions and maximum population size of $6$. After each run, the heuristic presented in Alg.~\ref{alg:overlapheuristic} was applied to the best solution returned by cob-aiNet[C].

The results were evaluated by the average of the Normalized Mutual Information, which indicates how close a given partition of the network is from the real partition (ground truth)~\cite{lancichinettiEtAl20011,olivetti2013identifying}. In this work, the solutions with non-overlapping (labeled NMI in the tables that follow) and overlapping (labeled NMI OVER.) partitions were evaluated. Besides, the obtained results for overlapping partitions were also compared to those obtained with the technique proposed in~\cite{olivetti2013identifying} (labeled NMI MULTIMODAL).

In the following tables (Tables~\ref{tab:networks} to~\ref{tab:realnetworks}), it is also reported the evaluated fitness of the returned solutions (labeled FIT), the number of overlapping communities obtained with the proposed heuristic (labeled \# Over.) and with the technique described in~\cite{olivetti2013identifying} (labeled \# Over. Multimodal), the number of communities found (labeled \# Comm.) and the total time in seconds taken (labeled TIME)\footnote{All the experiments were performed on an Intel Core i5 with 2.7GHz, 8GB of RAM and OSX 10.9.2.}. 


All the parameters required by Flex were empirically defined here for groups of networks with similar structures, and are presented in Table~\ref{tab:parameters}. Notice though that, in practice, these parameters should be set depending on the goal of the network analysis (e.g. if partitions with highly clustered communities are required, $\alpha$ should be set with higher values).

Also, if the partition structure of the network is known \textit{a priori}, the calibration of such parameters in order to obtain the known partition could indicate some characteristics of the network dynamics, such as, for example, the way that the connections of each node were established.

\begin{table}[t]
\caption{Weight parameters and heuristic thresholds for each dataset.}
\vspace{-5pt}
\centering{
\scriptsize{
\begin{tabular}{ccccccc}
\hline
Network & $\alpha$ & $\beta$ & $\gamma$ & $thr\bigtriangleup$ & $thrN$ & $thrSh$ \\
\hline
Network 50, Karate & $0.8$ & $0.3$ & $2$ & $0.3$ & $0.6$ & $0.25$ \\
Network 100-500, Krebs & $0.8$ & $0.3$ & $4$ & $0.3$ & $0.7$ & $0.25$ \\
Noise Network 50-500 & $0.5$ & $1.0$ & $4$ & $0.3$ & $0.7$ & $0.45$ \\
Football & $0.8$ & $0.6$ & $4$ & $0.3$ & $0.6$ & $0.25$ \\
Dolphins & $0.4$ & $0.3$ & $4$ & $0.3$ & $0.6$ & $0.25$ \\
\hline
\end{tabular}
}
}
\label{tab:parameters}
\end{table}

\subsection{Artificial Networks with Overlapping Communities}
\label{sec:artnetworks}

As expected, for the first set of networks (results given in Tab.~\ref{tab:networks}), both Flex and Modularity obtained the same values of NMI for every experiment. This is due to the lower rate of noise adopted in the creation of such networks, which makes the identification of the real partitions trivial for most fitness functions. Notice though that the overlapping detection heuristic proposed here resulted in partitions with perfect NMI score (equal to 1.0) for every network considered, except Network 100. In this particular dataset, one of the overlapping nodes did not attend the criteria established by the heuristic, thus the obtained NMI was slightly lower than 1.0.

\begin{table}[t]
\caption{Results for the artificial networks with overlap and low level of noise.}
\vspace{-18pt}
\centering{
\scriptsize{
\subfloat[Network 50]{\input{network50.tex}}
\subfloat[Network 100]{\input{network100.tex}} \\
\subfloat[Network 200]{\input{network200.tex}}
\subfloat[Network 500]{\input{network500.tex}}
}
}
\label{tab:networks}
\vspace{-15pt}
\end{table}

The results for the second set of networks, which were generated with higher noise, are reported in Tab.~\ref{tab:noisenetworks}. In this scenario, the differences between Flex and Modularity become much more evident. In every situation the heuristic combined with Flex was able to find most of the overlapping nodes of each problem, as pointed out by the higher values of NMI. On the other hand, the method proposed in~\cite{olivetti2013identifying} obtained lower values of NMI by introducing many more (false) overlapping nodes into the partition. It is also noticeable that, in the presence of noise, Modularity also tends to find partitions with much less communities than Flex, which is due to the resolution limit discussed in Sect.~\ref{sec:flex}.

\begin{table}[t]
\caption{Results for the artificial networks with overlap and high level of noise.}
\vspace{-18pt}
\centering{
\scriptsize{
\subfloat[Noise Network 50]{\input{network50_03.tex}}
\subfloat[Noise Network 100]{\input{network100_03.tex}} \\
\subfloat[Noise Network 200]{\input{network200_03.tex}}
\subfloat[Noise Network 500]{\input{network500_03.tex}}
}}
\label{tab:noisenetworks}
\vspace{-15pt}
\end{table}

It is important to highlight that the results were statistically verified by means of the Kruskal paired test with significance $< 0.05$. Therefore, in Tables~\ref{tab:networks}--\ref{tab:realnetworks}, those results with statistical significance (p-value$<0.05$) are marked in bold.

\vspace{-5pt}
\subsection{Real-world Social Networks}
\label{sec:realworld}
\vspace{-5pt}

Regarding the real-world social networks, the obtained results (given in Tab.~\ref{tab:realnetworks}) show that, again, Flex leads to a significant improvement over Modularity. However, it is important to notice that the ground truths of such networks are related to the classification of the nodes of these datasets according to their respective domains, so it does not necessarily mean that they actually correspond to the true partitions of the networks. Therefore, it maybe practically impossible to reach a perfect NMI score in some situations. It is also important to notice that those ground truth were originally devised without overlapping, so the NMI score with overlapping will always be smaller than the original NMI score.

Some interesting characteristics of each of these networks can be identified through a combination of visual inspection of the obtained partitions together with an analysis of the weights ($\alpha$ and $\beta$) that led to the best values of NMI. From the obtained results, it is possible to infer that both the Karate Club and Krebs networks are formed by highly clustered communities ($\alpha=0.8$), which makes sense as the Karate Club is a small social network prone to mutual friendships and the Krebs network, captures the interest of readers about particular subjects, so they tend to buy only books that are related to their political views.

The Football network required the same value for $\alpha$ but a much higher value for $\beta$, which means that this particular network does not allow open triangles. The reason for that is due to the organization of tournaments that limit the occurrence of intra-cluster relationships. Finally, for the Dolphins network the required weights are more favorable to the establishment of inter-community relationships instead of clustering, which might be related to the hierarchy in their society that favors the creation of several hubs inside a community~\cite{Lusseau2003}, thus raising the number of open triangles.

\begin{table}[t]
\caption{Results obtained for the real world networks.}
\vspace{-18pt}
\centering{
\scriptsize{
\subfloat[Karate Club]{\input{karate.tex}}
\subfloat[Dolphins]{\input{dolphins.tex}} \\
\subfloat[Football]{\input{footballTSEinput.tex}}
\subfloat[Krebs]{\input{krebs.tex}}
}}
\label{tab:realnetworks}
\vspace{-18pt}
\end{table}

\begin{figure}[t]
\subfloat[Karate Club]{\includegraphics[trim=0cm 12cm 0cm 1cm, clip=true,width=0.42\textwidth]{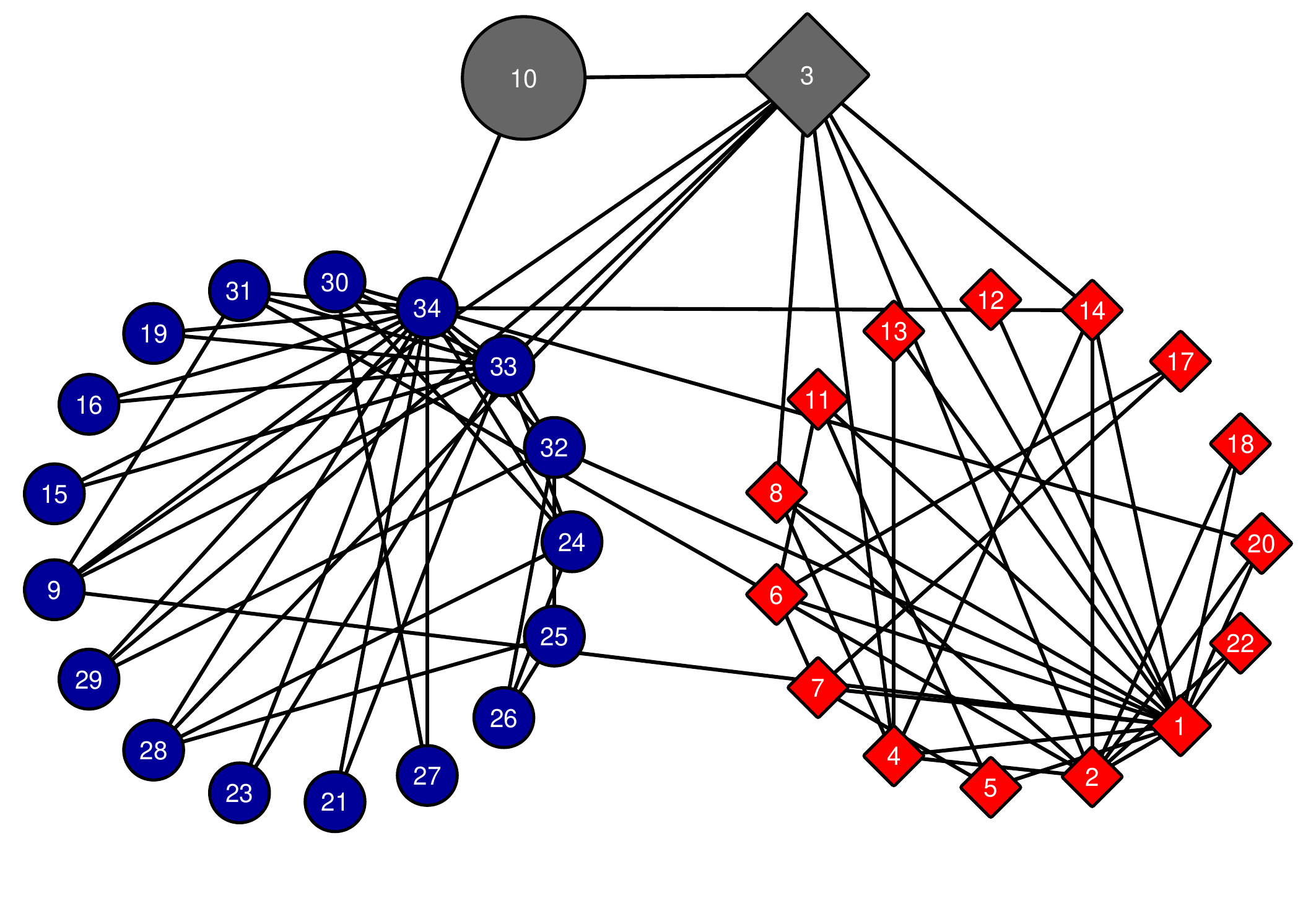}}
\subfloat[Dolphins]{\includegraphics[trim=0cm 20cm 8cm 0cm, clip=true,width=0.5\textwidth]{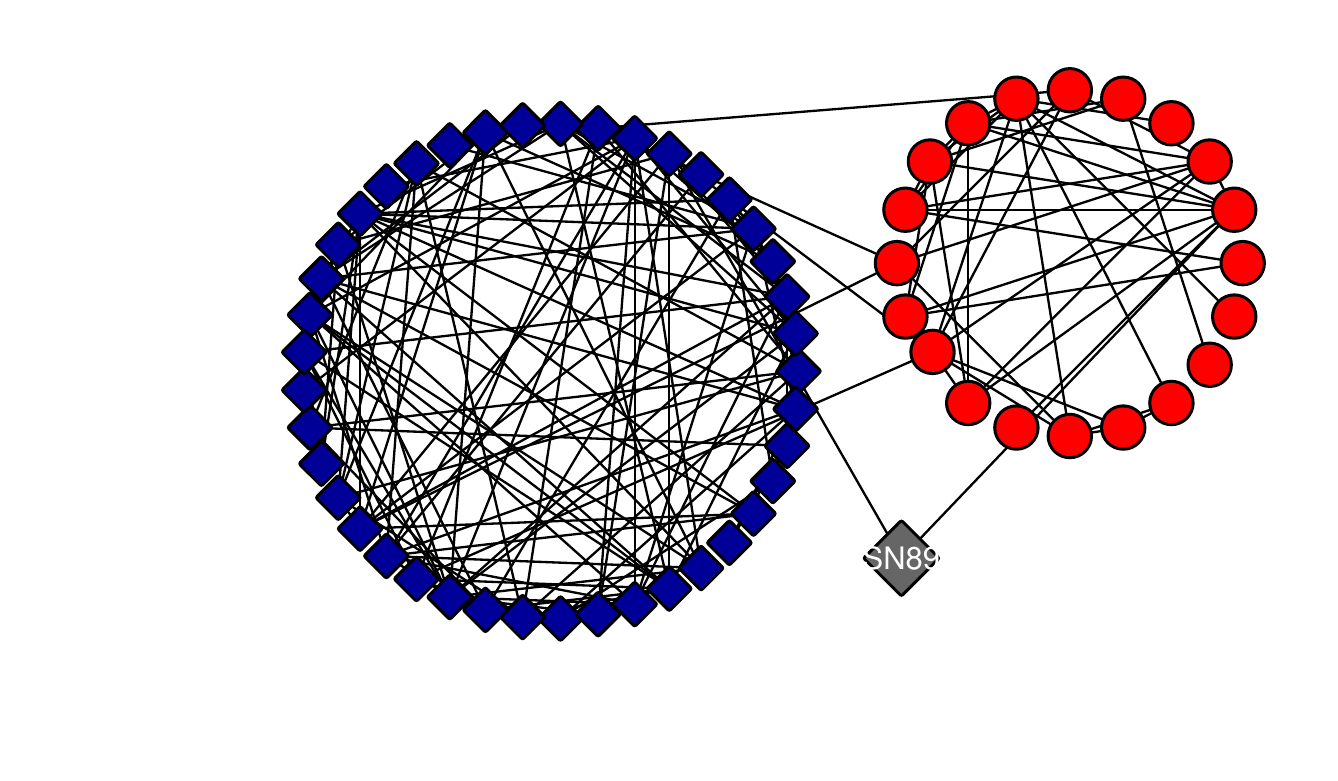}}
\\
\subfloat[Football]{\includegraphics[trim=0cm 14cm 4cm 1cm, clip=true,width=0.5\textwidth]{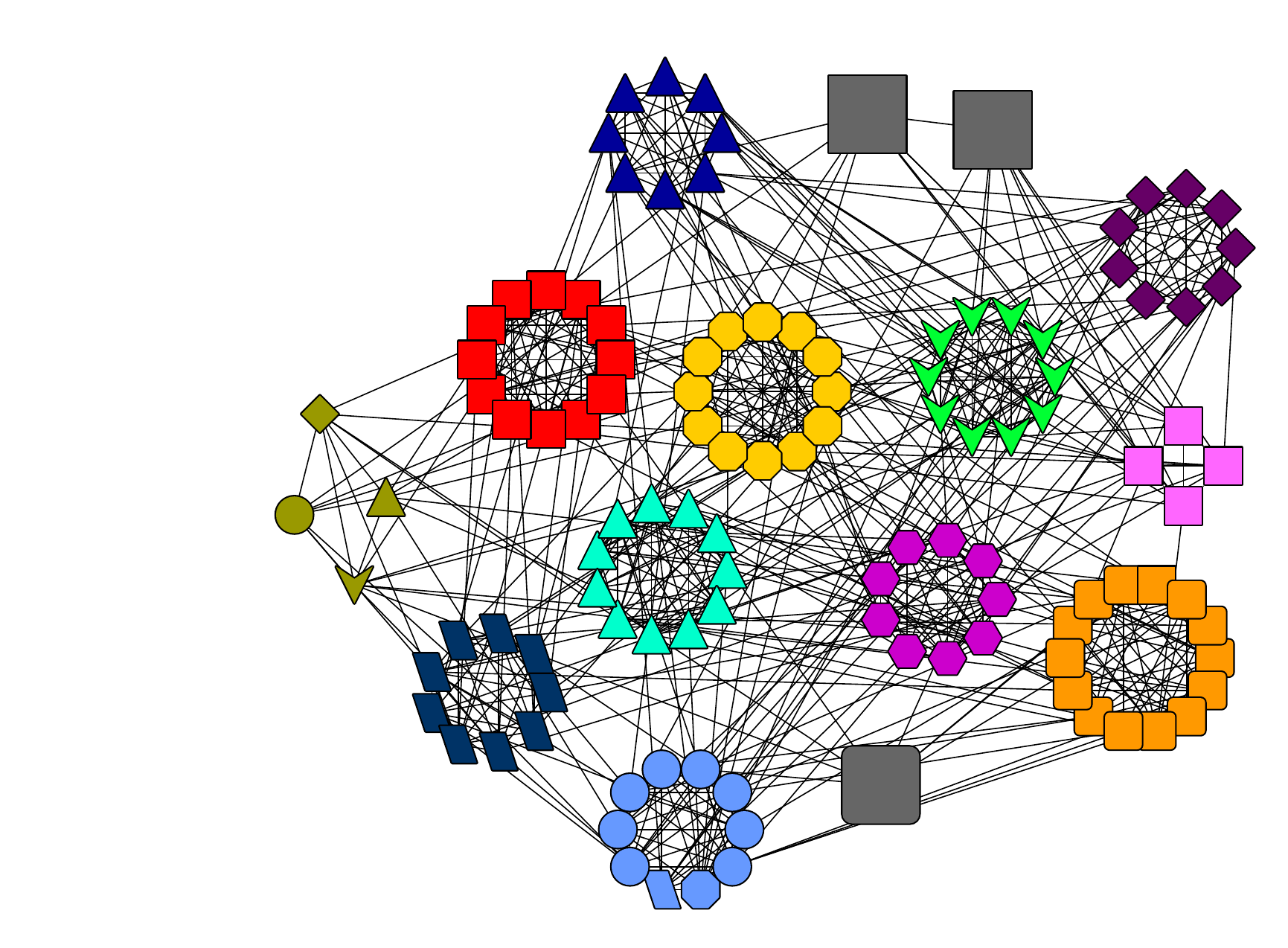}}
\subfloat[Krebs]{\includegraphics[trim=2cm 20cm 10cm 0cm, clip=true,width=0.45\textwidth]{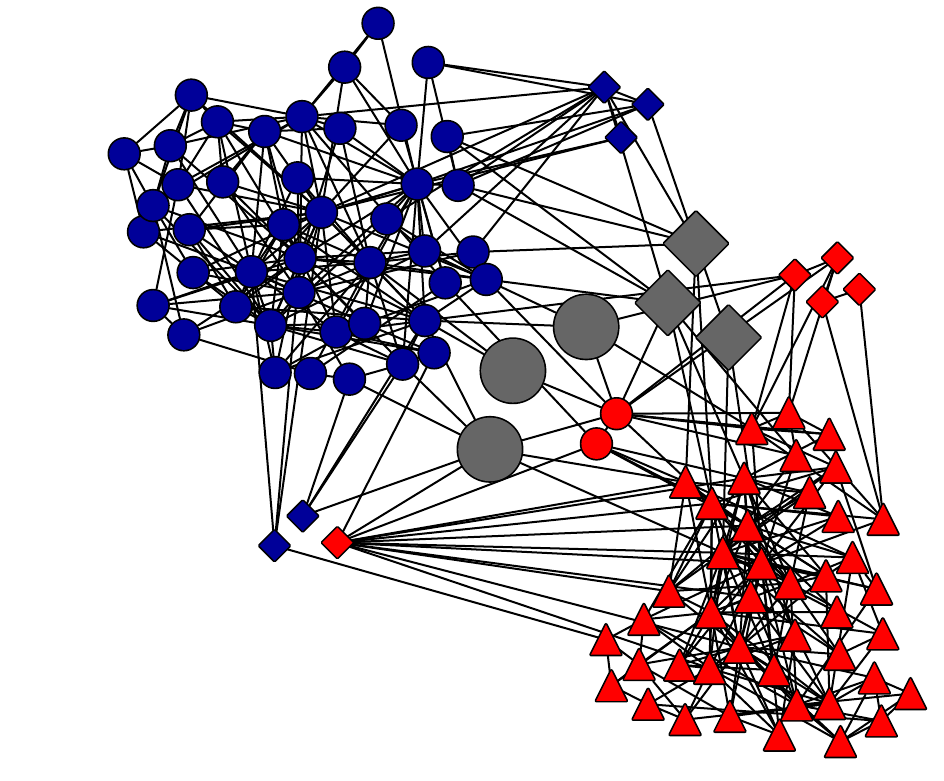}}
\caption{Results obtained by cob-aiNet[C] with Flex for the real-world networks.}
\label{fig:realnetworks}
\vspace{-8pt}
\end{figure}

In order to illustrate the overlapping communities obtained by the combination of Flex and the proposed heuristic, Fig.~\ref{fig:realnetworks} depicts the best partitions with overlapping nodes for each problem. In Fig.~\ref{fig:realnetworks}, the colors represent the communities found by the optimization of Flex, the shapes represent the communities in the ground truth and the larger nodes are the overlapping nodes. It is visually noticeable that the community formation found using the Flex function makes sense, given the weights for each network. Also, every overlapping node is clearly positioned between two or more distinct groups.

\vspace{-5pt}
\section{Conclusion}
\label{sec:conclusion}
\vspace{-5pt}

In this paper a novel fitness function for community detection in complex networks was introduced, together with a heuristic that, based on particular characteristics of this function, allows the identification of overlapping nodes. This new fitness function, called Flex, is parametrized in such a way that it can be adapted to obtain communities with different characteristics. 

Through an extensive experimental setup, in which Flex was combined with an immune-inspired algorithm adopted for the optimization process (a novel mutation operator was also proposed in this paper), it was possible to verify that this new fitness function and heuristic are capable of leading to partitions close to the ground truth of a set of networks with different characteristics.

As for future investigations, we intend to evaluate a multi-objective version of the proposed fitness function and verify whether it leads to a better set of overlapping nodes. Besides, a thorough sensitivity analysis of Flex's parameters will also be performed.

%
%

\bibliographystyle{splncs}
\bibliography{FlexibleFitnessCommunityDetection}

\end{document}

%% file: network50.tex
	\begin{tabular}{lll}
& Flex & Modularity \\\hline
FIT: & $0.81$ & $0.60$ \\
NMI: & $0.94$ & $0.94$ \\
NMI OVER.: & $1.00$ & $1.00$ \\
NMI MULTIMODAL: & $0.87$ & $\mathbf{0.93}$ \\
\# Comm.: & $5.00$ & $5.00$ \\
\# Over.: & $3.00$ & $3.00$ \\
\# Over. Multimodal: & $4.50$ & $0.95$ \\
TIME: & $48.52$ & $32.73$ \\
\hline
	\end{tabular}

%% file: network100.tex
	\begin{tabular}{lll}
& Flex & Modularity \\\hline
FIT: & $0.81$ & $0.59$ \\
NMI: & $0.95$ & $0.95$ \\
NMI OVER.: & $0.98$ & $0.98$ \\
NMI MULTIMODAL: & $\mathbf{0.96}$ & $0.95$ \\
\# Comm.: & $4.00$ & $4.00$ \\
\# Over.: & $2.10$ & $2.00$ \\
\# Over. Multimodal: & $1.05$ & $0.50$ \\
TIME: & $224.33$ & $146.19$ \\
\hline
	\end{tabular}

%% file: network200.tex
	\begin{tabular}{lll}
& Flex & Modularity \\\hline
FIT: & $0.85$ & $0.64$ \\
NMI: & $0.97$ & $0.97$ \\
NMI OVER.: & $1.00$ & $1.00$ \\
NMI MULTIMODAL: & $\mathbf{0.98}$ & $0.97$ \\
\# Comm.: & $5.00$ & $5.00$ \\
\# Over.: & $3.00$ & $3.00$ \\
\# Over. Multimodal: & $0.65$ & $0.00$ \\
TIME: & $406.03$ & $113.93$ \\
\hline
	\end{tabular}

%% file: network500.tex
	\begin{tabular}{lll}
& Flex & Modularity \\\hline
FIT: & $0.87$ & $0.80$ \\
NMI: & $0.99$ & $0.99$ \\
NMI OVER.: & $1.00$ & $1.00$ \\
NMI MULTIMODAL: & $0.99$ & $0.99$ \\
\# Comm.: & $11.95$ & $12.00$ \\
\# Over.: & $2.95$ & $3.00$ \\
\# Over. Multimodal: & $0.00$ & $0.00$ \\
TIME: & $590.99$ & $332.31$ \\
\hline
	\end{tabular}

%% file: network50_03.tex
	\begin{tabular}{lll}
& Flex & Modularity \\\hline
FIT: & $0.47$ & $0.55$ \\
NMI: & $\mathbf{0.77}$ & $0.43$ \\
NMI OVER.: & $\mathbf{0.78}$ & $0.43$ \\
NMI MULTIMODAL: & $\mathbf{0.69}$ & $0.42$ \\
\# Comm.: & $6.00$ & $3.00$ \\
\# Over.: & $3.00$ & $0.00$ \\
\# Over. Multimodal: & $12.80$ & $1.60$ \\
TIME: & $58.77$ & $73.34$ \\
\hline
	\end{tabular}

%% file: network100_03.tex
	\begin{tabular}{lll}
& Flex & Modularity \\\hline
FIT: & $0.51$ & $0.49$ \\
NMI: & $\mathbf{0.93}$ & $0.58$ \\
NMI OVER.: & $\mathbf{0.94}$ & $0.58$ \\
NMI MULTIMODAL: & $\mathbf{0.82}$ & $0.43$ \\
\# Comm.: & $6.00$ & $4.30$ \\
\# Over.: & $0.40$ & $1.45$ \\
\# Over. Multimodal: & $15.60$ & $35.40$ \\
TIME: & $143.46$ & $200.52$ \\
\hline
	\end{tabular}

%% file: network200_03.tex
	\begin{tabular}{lll}
& Flex & Modularity \\\hline
FIT: & $0.51$ & $0.50$ \\
NMI: & $\mathbf{0.93}$ & $0.60$ \\
NMI OVER.: & $\mathbf{0.94}$ & $0.62$ \\
NMI MULTIMODAL: & $\mathbf{0.81}$ & $0.47$ \\
\# Comm.: & $6.00$ & $4.00$ \\
\# Over.: & $2.00$ & $5.50$ \\
\# Over. Multimodal: & $34.65$ & $40.55$ \\
TIME: & $655.99$ & $657.93$ \\
\hline
	\end{tabular}

%% file: network500_03.tex
	\begin{tabular}{lll}
& Flex & Modularity \\\hline
FIT: & $0.50$ & $0.63$ \\
NMI: & $\mathbf{0.89}$ & $0.55$ \\
NMI OVER.: & $\mathbf{0.91}$ & $0.56$ \\
NMI MULTIMODAL: & $\mathbf{0.73}$ & $0.40$ \\
\# Comm.: & $13.95$ & $7.65$ \\
\# Over.: & $10.30$ & $6.70$ \\
\# Over. Multimodal: & $145.70$ & $194.00$ \\
TIME: & $1818.17$ & $2619.40$ \\
\hline
	\end{tabular}

%% file: karate.tex
	\begin{tabular}{lll}
& Flex & Modularity \\\hline
FIT: & $0.82$ & $0.41$ \\
NMI: & $\mathbf{0.95}$ & $0.40$ \\
NMI OVER.: & $\mathbf{0.79}$ & $0.45$ \\
NMI MULTIMODAL: & $\mathbf{0.91}$ & $0.55$ \\
\# Comm.: & $1.95$ & $3.95$ \\
\# Over.: & $1.90$ & $2.90$ \\
\# Over. Multimodal: & $0.55$ & $16.30$ \\
TIME: & $39.36$ & $50.24$ \\
\hline
	\end{tabular}

%% file: dolphins.tex
	\begin{tabular}{lll}
& Flex & Modularity \\\hline
FIT: & $0.68$ & $0.53$ \\
NMI: & $\mathbf{0.86}$ & $0.46$ \\
NMI OVER.: & $\mathbf{0.82}$ & $0.46$ \\
NMI MULTIMODAL: & $\mathbf{0.82}$ & $0.46$ \\
\# Comm.: & $2.30$ & $4.00$ \\
\# Over.: & $1.90$ & $7.00$ \\
\# Over. Multimodal: & $4.60$ & $0.00$ \\
TIME: & $63.88$ & $27.61$ \\
\hline
	\end{tabular}

%% file: footballTSEinput.tex
	\begin{tabular}{lll}
& Flex & Modularity \\\hline
FIT: & $0.77$ & $0.60$ \\
NMI: & $\mathbf{0.74}$ & $0.67$ \\
NMI OVER.: & $\mathbf{0.72}$ & $0.66$ \\
NMI MULTIMODAL: & $\mathbf{0.74}$ & $0.67$ \\
\# Comm.: & $11.25$ & $9.60$ \\
\# Over.: & $2.25$ & $2.05$ \\
\# Over. Multimodal: & $0.00$ & $0.00$ \\
TIME: & $23.36$ & $26.31$ \\
\hline
	\end{tabular}

%% file: krebs.tex
	\begin{tabular}{lll}
& Flex & Modularity \\\hline
FIT: & $0.72$ & $0.53$ \\
NMI: & $\mathbf{0.45}$ & $0.32$ \\
NMI OVER.: & $\mathbf{0.43}$ & $0.37$ \\
NMI MULTIMODAL: & $\mathbf{0.45}$ & $0.32$ \\
\# Comm.: & $2.25$ & $4.95$ \\
\# Over.: & $7.30$ & $11.90$ \\
\# Over. Multimodal: & $6.20$ & $0.75$ \\
TIME: & $126.01$ & $47.33$ \\
\hline
	\end{tabular}